\documentclass[10pt,twocolumn,letterpaper]{article}

\usepackage[pagenumbers]{cvpr} % To force page numbers, e.g. for an arXiv version

\usepackage{graphicx}
\usepackage{amsmath}
\usepackage{amssymb}
\usepackage{booktabs}

\usepackage{color}

\usepackage{multirow}

\usepackage{pifont}% http://ctan.org/pkg/pifont

\usepackage[normalem]{ulem}

\newcommand{\cmark}{\ding{51}}%
\newcommand{\xmark}{\ding{55}}%

\usepackage[pagebackref,breaklinks,colorlinks]{hyperref}

\usepackage[accsupp]{axessibility}

\usepackage[capitalize]{cleveref}
\crefname{section}{Sec.}{Secs.}
\Crefname{section}{Section}{Sections}
\Crefname{table}{Table}{Tables}
\crefname{table}{Tab.}{Tabs.}

\def\cvprPaperID{3366} % *** Enter the CVPR Paper ID here
\def\confName{CVPR}
\def\confYear{2023}

\newif\ifcomments
\commentsfalse %Remove comments

\newcommand{\commentZ}[1]{\ifcomments{\color{red}XZY: \bf\color{red}#1\color{black}}\fi}

\newcommand{\xd}[1]{\ifcomments{\color{blue}XD: \bf\color{blue}#1\color{black}}\fi}

\definecolor{green}{rgb}{0,0.9,0.1}
\newcommand{\commentH}[1]{\ifcomments{\color{green}H: \bf\color{green}#1\color{black}}\fi}

\begin{document}

\title{GP-VTON: Towards General Purpose Virtual Try-on via \\ Collaborative Local-Flow Global-Parsing Learning\vspace{-2mm}}

\author{
Zhenyu Xie{$^{1}$},
~ Zaiyu Huang{$^{1}$},
~ Xin Dong{$^{2}$},
~ Fuwei Zhao{$^{1}$}\\ \vspace{-16pt}\\ 
Haoye Dong{$^{3}$},
~ Xijin Zhang{$^{2}$},
~ Feida Zhu{$^{2}$},
~ Xiaodan Liang{$^{1,4}$}\\\vspace{-10pt}\\
{$^{1}$}Shenzhen Campus of Sun Yat-Sen University, {$^{2}$}ByteDance \\ \vspace{-16pt}\\
{$^{3}$}Carnegie Mellon University,
{$^{4}$}Peng Cheng Laboratory\\
\small{\tt{\{xiezhy6,huangzy225,zhaofw\}@mail2.sysu.edu.cn,donghaoye@cmu.edu}}\\ 
\small{\tt{\{dongxin.1016,zhangxijin\}@bytedance.com,zhufeida@connect.hku.hk},xdliang328@gmail.com}
\vspace{-2mm}
}

% \and
% \maketitle

\twocolumn[{%
\renewcommand\twocolumn[1][]{#1}%
\maketitle
    \includegraphics[width=1.0\hsize]{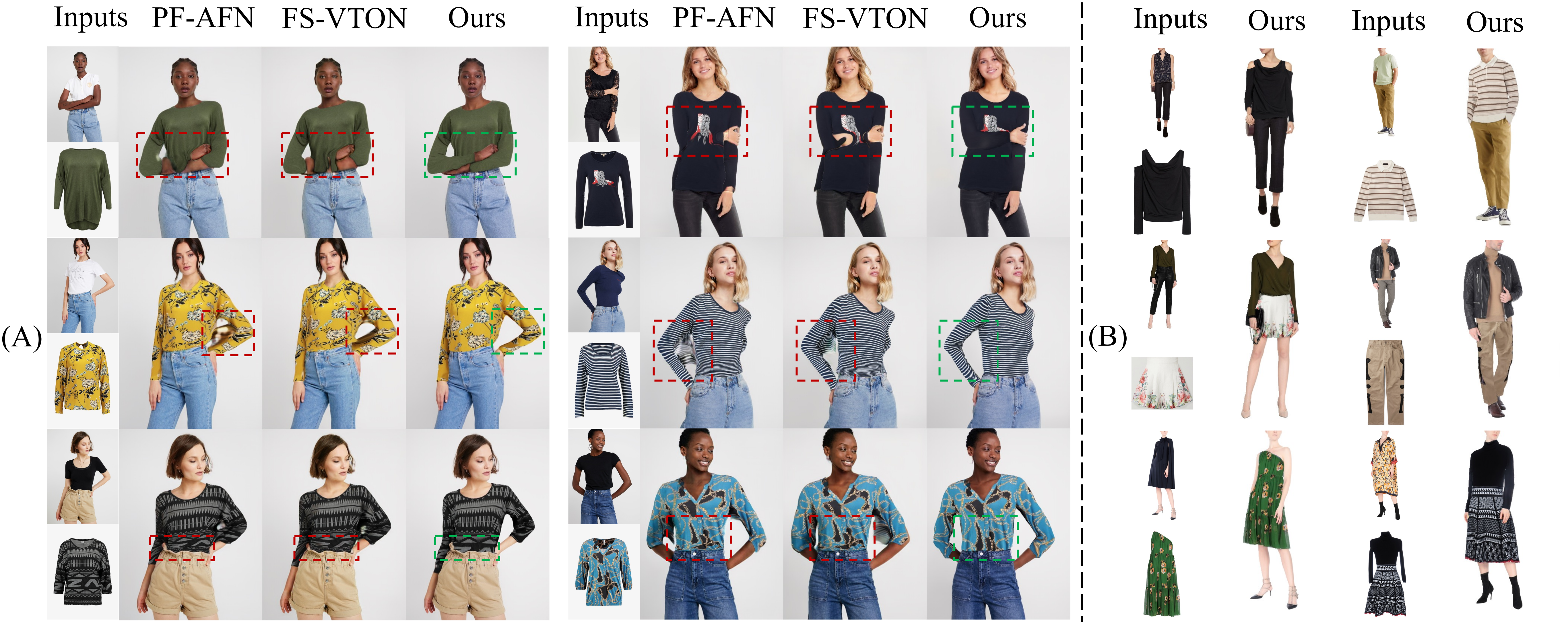}
    \vspace{-6mm}
    \captionof{figure}{
    % The results of our proposed method on the challenging try-on scenarios (left) and the multi-category try-on scenarios (right). 
    Our method (A) outperforms existing SOTA methods (e.g., PF-AFN~\cite{Ge2021PFAFN}, FS-VTON~\cite{he2022fs_vton}) on the challenging try-on scenario, and (B) can be easily extended to multi-category scenario to generate high-resolution photo-realistic try-on results.
    % On the left, when receiving challenging inputs (e.g., intricate pose, complex garment) our method outperforms existing SOTA methods (e.g., PF-AFN~\cite{Ge2021PFAFN}, FS-VTON~\cite{he2022fs_vton}), and is capable of  generating semantic-correct and distortion-free try-on results. On the right, our method can be easily extended to multi-category scenarios and generate high-resolution photo-realistic results.
    \xd{remove A,B to save space}} 
    \vspace{3mm}
    \label{fig:teaser}
}]

%%%%%%%%% ABSTRACT
\begin{abstract}
Image-based Virtual Try-ON aims to transfer an in-shop garment onto a specific person. Existing methods employ a global warping module to model the anisotropic deformation for different garment parts, which fails to preserve the semantic information of different parts when receiving challenging inputs (e.g, intricate human poses, difficult garments). Moreover, most of them directly warp the input garment to align with the boundary of the preserved region, which usually requires texture squeezing to meet the boundary shape constraint and thus leads to texture distortion. The above inferior performance hinders existing methods from real-world applications. To address these problems and take a step towards real-world virtual try-on, we propose a General-Purpose Virtual Try-ON framework, named GP-VTON, by developing an innovative Local-Flow Global-Parsing (LFGP) warping module and a Dynamic Gradient Truncation (DGT) training strategy. Specifically, compared with the previous global warping mechanism, LFGP employs local flows to warp garments parts individually, and assembles the local warped results via the global garment parsing, resulting in reasonable warped parts and a semantic-correct intact garment even with challenging inputs.On the other hand, our DGT training strategy dynamically truncates the gradient in the overlap area and the warped garment is no more required to meet the boundary constraint, which effectively avoids the texture squeezing problem. Furthermore, our GP-VTON can be easily extended to multi-category scenario and jointly trained by using data from different garment categories. Extensive experiments on two high-resolution benchmarks demonstrate our superiority over the existing state-of-the-art methods.~\footnote{Corresponding author is Xiaodan Liang. Code is available at ~\href{https://github.com/xiezhy6/GP-VTON}{gp-vton}.}

\commentH{Image-based Virtual Try-ON aims to seamlessly transfer an in-shop garment onto a specific person while preserving the cloth detail and human identity. Existing methods employ a global warping module to model the anisotropic deformation for different garment parts, which often fail to deform different cloth parts correctly by leveraging the semantic information when receiving challenging inputs (e.g, intricate human poses, difficult garments). They also heavily rely on warp boundary given by preserved region, which leads to texture distortion at boundary area. In this paper, we propose GP-VTON, a a General-Purpose Virtual Try-ON framework to address the above problems. Our GP-VTON develops an innovative Local-Flow Global-Parsing(LFGP) warping module to strengthen the semantic awareness of flow prediction  and a novel Dynamic Gradient Truncation(DGT) training strategy to tackle boundary distortion. Specifically,  our LFGP warping module deforms each garment part via the local flow field individually, and assembles local warped results via the global garment parsing, resulting in reasonable warped parts and a semantic-correct intact garment even with challenging inputs. On the other hand, our DGT training strategy dynamically truncate the gradient in the overlap between the warped garment and the preserved region, which effectively alleviate the texture distortion at the boundary of the warped garment. Furthermore, since we explicitly divide and process different cloth parts, our GP-VTON can be easily extended for various garment categories and jointly trained by using data from different categories, leading to a unified framework for multi-category virtual try-on. Extensive experiments on two high-resolution benchmarks demonstrate our superiority over the existing state-of-the-art methods. Our code and model will be released upon acceptance.}
\end{abstract}

%%%%%%%%% BODY TEXT
\section{Introduction}
\label{sec:intro}
The problem of Virtual Try-ON (VTON), aiming to transfer a garment onto a specific person, is of particular importance for the nowadays e-commerce and the future metaverse. 
Compared with the 3D-based solutions~\cite{peng2012drape,fabian2014scs,gerard2017clothcap,lahner2018deepwrinkles,bhatnagar2019multi} which rely upon 3D scanning equipment or labor-intensive 3D annotations, the 2D image-based methods~\cite{bochao2018cpvton,xintong2019clothflow,han2020acgpn,VTON_zhao2021m3d,Ge2021PFAFN,kedan2021ovnet,zhenyu2021wasvton,zhenyu2021pastagan,he2022fs_vton,dong2022wflow,zaiyu20223dgcl,morelli2022dresscode,lee2022hrviton,bai2022sdafn}, which directly manipulate on the images, are more practical for the real world scenarios and thus have been intensively explored in the past few years. 

Although the pioneering 2D image-based VTON methods~\cite{Ge2021PFAFN,he2022fs_vton,lee2022hrviton} can synthesize compelling results on the widely used benchmarks~\cite{xintong2018viton,dong2019mgvton,choi2021vtonhd}, there still exist some deficiencies preventing them from the real-world scenarios, which we argue mainly contain three-folds.
First, existing methods have strict constraints on the input images, and are prone to generate artifacts when receiving challenging inputs.
% To be specific, as shown in the first row of Fig.~\ref{fig:teaser}(A), 
To be specific, as shown in the 1st row of Fig.~\ref{fig:teaser}(A), 
when the pose of the input person is intricate, existing methods~\cite{Ge2021PFAFN,he2022fs_vton} fail to preserve the semantic information of different garment parts, resulting in the indistinguishable warped sleeves. Besides, as shown in the 2nd row of Fig.~\ref{fig:teaser}(A), 
% for the existing methods~\cite{Ge2021PFAFN,he2022fs_vton}, 
if the input garment is a long sleeve without obvious seam between the sleeve and torso, existing methods will generate adhesive artifact between the sleeve and torso.
% there will be an adhesive region between the sleeve and torso in the result.
\commentH{Besides, as shown in the second row of Fig.~\ref{fig:teaser}(A), if the input garment is a long sleeve without obvious seam between the sleeve and torso, there will often be an adhesive region between the sleeve and torso in the try-on result.}
Second, most of the existing methods directly squeeze the input garment to make it align with the preserved region, leading to the distorted texture around the preserved region (e.g., the 3rd row of Fig.~\ref{fig:teaser}(A)).
Third, most of the existing works only focus on the upper-body try-on and neglect other garment categories (i.e, lower-body, dresses), which further limits their scalability for real-world scenarios.

To relieve the input constraint for VTON systems and fully 
exploit their application potential, in this paper, we take a step forwards and propose a unified framework, named GP-VTON, for the \textbf{G}eneral-\textbf{P}urposed \textbf{V}irtual \textbf{T}ry-\textbf{ON}, which can generate realistic try-on results even for the challenging scenario (Fig.~\ref{fig:teaser}(A)) (e.g., intricate human poses, difficult garment inputs, etc.), and can be easily extended to the multi-category scenario (Fig.~\ref{fig:teaser}(B)).

\commentH{First, existing methods are usually not generalizable to  challenging inputs of garment and person images due to inefficency of capturing semantic information.
To be specific, as shown in the first row of Fig.~~\ref{}(a), when the pose of the input person is intricate, existing methods~\cite{Ge2021PFAFN,he2022fs_vton} fail to preserve the semantic identities of different garment parts during the try-on process, resulting in the indistinguishable warped sleeves. The second row of Fig.~~\ref{}(a) shows another example, for the existing methods~\cite{Ge2021PFAFN,he2022fs_vton}, if the input garment is a long sleeve without obvious seam between the sleeve and torso, there will be an adhesive region between the sleeve and torso in the try-on result.
Second, most of the existing works, with a single global deformation process, only focus on the upper-body virtual try-on, which are unable to adapt into more practical real-world scenarios containing various garment types (i.e, lower-body, dresses).
Our newly proposed GP-VTON, the \textbf{G}eneral-\textbf{P}urposed \textbf{V}irtual \textbf{T}ry-\textbf{ON}, however, aims to build up an unified virtual try-on framework which can not only support garment transfer for diverse garment categories ~\ref{}(b)(e.g., upper, lower, dresses),  but also synthesize realistic try-on results in some challenging scenarios~\ref{}(a) (e.g., intricate human poses, difficult garment inputs, etc.)}

The innovations of our GP-VTON lie in a novel Local-Flow Global-Parsing (LFGP) warping module and a Dynamic Gradient Truncation (DGT) training strategy for the warping network, which enable the network to generate high fidelity deformed garments, \commentH{high fidelity deformed garments} and further facilitate our GP-VTON to generate photo-realistic try-on results.
% for diverse VTON scenarios even with challenging inputs. 

\commentH{The innovations of our GP-VTON lie in a novel Local-Flow-Global-Parsing (LFGP) warping module and a Dynamic Gradient Truncation (DGT) training strategy for the warping network, which enables the network to achieve semantic-correct and distortion-free garments deformation, and further facilitate our GP-VTON to generate photo-realistic results for diverse VTON scenarios even with challenging inputs. }
\xd{too long paragraph, split into two}
Specifically, most of the existing methods employ neural network to model garment deformation by introducing the Thin Plate Splines (TPS) transformation~\cite{bookstein1989tps} or the appearance flow~\cite{TinghuiZhou2016ViewSB} into the network, and training the network in a weakly supervised manner (i.e., without ground truth for the deformation function). However, both of the TPS-based methods~\cite{xintong2018viton,bochao2018cpvton,han2020acgpn,choi2021vtonhd,kedan2021ovnet} and the flow-based methods~\cite{xintong2019clothflow,Ge2021PFAFN,he2022fs_vton,lee2022hrviton,bai2022sdafn} directly learn a global deformation field, therefore fail to represent complicated non-rigid garment deformation that requires diverse transformation for different garment parts. Taking the intricate pose case in Fig.~\ref{fig:teaser}(A) as an example, existing methods~\cite{Ge2021PFAFN,he2022fs_vton} can not simultaneously guarantee accurate deformation for the torso region and sleeve region, and lead to exceeding distorted sleeves.
In contrast, 
% inspired by the divide-and-conquer principle, 
our LFGP warping module chooses to learn diverse local deformation fields for different garment parts, which is capable of individually warping each garment part, and generating semantic-correct warped garment even for intricate pose case. Besides, since each local deformation field merely affects one corresponding garment part, garment texture from other parts is agnostic to the current deformation field and will not appear in the current local warped result. Therefore, the garment adhesion problem in the complex garment scenario can be completely addressed (as demonstrated in the 2nd row of Fig.~\ref{fig:teaser}(A)). However, directly assembling the local warped parts together can not obtain realistic warped garments, because there would be overlap among different warped parts. To deal with this, our LFGP warping module collaboratively estimates a global garment parsing to fuse different local warped parts, resulting in a complete and unambiguous warped garment.

\commentH{Specifically, compared to most of the existing TPS-based ~\cite{xintong2018viton,bochao2018cpvton,han2020acgpn,choi2021vtonhd,kedan2021ovnet} and the flow-based~\cite{xintong2019clothflow,Ge2021PFAFN,he2022fs_vton,lee2022hrviton,bai2022sdafn} VTON methods, which simply learn a global deformation field to represent complicated non-rigid garment deformation, our LFGP warping module chooses to learn diverse local deformation fields for different garment parts following a divide-and-conquer principle, which is capable of individually warping each garment part, and generating semantic-correct warped garment even for intricate pose cases. Besides, since each local deformation field merely affects one corresponding garment part, garment texture from other parts is agnostic to the current deformation field and will not appear in the current local warped result. Therefore, the garment adhesion problem in the hard garment input scenario can be completely addressed (as demonstrated in the second row of Fig.~\ref{fig:teaser}(A)). However, directly assembling the local warped parts together can not obtain realistic warped garments since there would be overlap among different warped parts. To deal with this, our LFGP warping module collaboratively estimates a global garment parsing to fuse different local warped parts, resulting in a complete and unambiguous warped garment.}

\begin{figure*}
\begin{center}
\includegraphics[width=1.0\hsize]{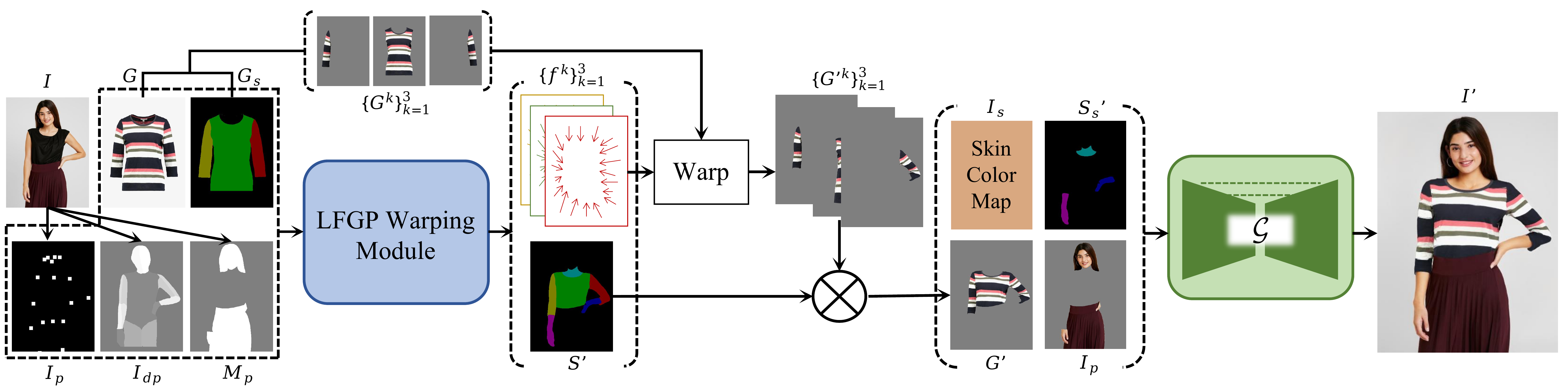}
\end{center}
    \vspace{-6mm}
   \caption{Overview of GP-VTON. The LFGP warping module aims to estimate the local flows $\{f^k\}_{k=1}^3$ and global garment parsing $S'$, which is used to warp different garment parts $\{G^k\}_{k=1}^3$ and assembles warped parts $\{G'^k\}_{k=1}^3$ into the intact garment $G'$, respectively. The generator $\mathcal{G}$ takes as inputs $G'$ and the other person-related conditions to generate the final try-on result $I'$.  \xd{caption}}
  \vspace{-4mm}
\label{fig:framework}
\end{figure*}

On the other hand, the warping network in existing methods~\cite{Ge2021PFAFN,he2022fs_vton,lee2022hrviton} takes as inputs the flat garment and the mask of the preserved region (i.e., region to be preserved during the try-on procedure, such as the lower garment for upper-body VTON), and force the input garment to align with the boundary of the preserved region (e.g., the junction of the upper and lower garment), which usually require garment squeezing to meet the shape constraint and lead to texture distortion around the garment junction (please refer to the 3rd row of Fig.~~\ref{fig:teaser}(A)). An effective solution to this problem is exploiting the gradient truncation strategy for the network training, in which the warped garment will be 
processed by the preserved mask before calculating the warping loss and the gradient in the preserved region will not be back-propagated. By using such a strategy, the warped garment is no longer required to strictly align with the preserved boundary, which largely avoids the garment squeezing and texture distortion. 
However, due to the poor supervision of warped garments in the preserved region, directly employing the gradient truncation for all training data will lead to excessive freedom for the deformation field, which usually results in texture stretching in the warped results. \commentH{"To tackle this problem", or add a sentence describing the purpose? } To tackle this problem, we proposed a Dynamic Gradient Truncation (DGT) training strategy which dynamically conducts gradient truncation for different training samples according to the disparity of height-width ratio between the flat garment and the warped garment. By introducing the dynamic mechanism, our LFGP warping module can alleviate the texture stretching problem and obtain realistic warped garments with better texture preservation.

\commentH{On the other hand, based on the problem of texture distortion around the garment junction (as shown in the third row of Fig.~~\ref{fig:teaser}(A)) that commonly exists in recent methods~\cite{Ge2021PFAFN,he2022fs_vton,lee2022hrviton}, we introduce proposed a Dynamic Gradient Truncation (DGT) training strategy to 

the warping network in existing methods~\cite{Ge2021PFAFN,he2022fs_vton,lee2022hrviton} takes as inputs the flat garment and the mask of the preserved region (i.e., region to be preserved during the try-on procedure, such as the lower garment for upper-body VTON), and force the input garment to align with the boundary of the preserved region (e.g., the junction of the upper and lower garment), which usually require garment squeezing to meet the shape constraint and lead to texture distortion around the garment junction (please refer to the third row of Fig.~~\ref{fig:teaser}(A)). An effective solution to this problem is exploiting the gradient truncation strategy for the network training, in which the warped garment will be 
processed by the preserved mask before calculating the warping loss and the gradient in the preserved region will not be back-propagated. By using such a strategy, the warped garment is no longer required to strictly align with the preserved boundary, which largely avoids the garment squeezing and texture distortion. 
However, due to the poor supervision of warped garments in the preserved region, directly employing the gradient truncation for all training data will lead to excessive freedom for the deformation field, which usually results in texture stretching in the warped results. \commentH{"To tackle this problem", or add a sentence describing the purpose? } To tackle this problem, we proposed a Dynamic Gradient Truncation (DGT) training strategy which dynamically conducts gradient truncation for different training samples according to the disparity of height-width ratio between the flat garment and the warped garment. By introducing the dynamic mechanism, our LFGP warping module can alleviate the texture stretching problem and obtain realistic warped garments with better texture preservation.}

Overall, our contributions can be summarized as follows: (1) We propose a unified try-on framework, named GP-VTON, to generate photo-realistic results for diverse scenarios. (2) We propose a novel LFGP warping module to generate semantic-correct deformed garments even with challenging inputs. (3) We introduce a simple, yet effective DGT training strategy for the warping network to obtain distortion-free deformed garments. (4) Extensive experiments on two challenging high-resolution benchmarks show the superiority of GP-VTON over existing SOTAs.
% \footnote{{Our code and model will be released upon acceptance.}}

% \begin{itemize}
%     \item We propose a general-purposed virtual try-on framework, named GP-VTON, which can generate high resolution photo-realistic results in diverse scenarios.
%     \item We propose a novel LFGP warping algorithm to generate semantic-correct deformed garments even with challenging inputs.
%     \item We introduce a simple, yet effective DGT training strategy for the warping network to obtain distortion-free deformed garments.
%     \item Extensive experiments on two challenging high-resolution benchmarks show the superiority of GP-VTON over existing SOTAs.
% \end{itemize}

\section{Related Work}

\textbf{Human-centric Image Synthesis.}
Generative Adversarial Networks (GANs)~\cite{NIPS2014_5ca3e9b1}, especially the StyleGAN-based models~\cite{karras2019stylegan,karras2020stylegan2,karras2020stylegan2ada,karras2021stylegan3}, have recently achieved significant success for photo-realistic image synthesis. Therefore, in the field of human synthesis , most of the existing methods~\cite{anna2022insetgan,fu2022styleganhuman} inhrit the StyleGAN-based architecture to obtain high-fidelity synthesized results. InsetGAN~\cite{anna2022insetgan} combines the results from several pretrained GANs to a full-body human image, in which different pretrained GANs are in charge of the generation of different body parts (e.g., human body, face, hands, etc). StyleGAN-Human~\cite{fu2022styleganhuman} explores three crucial factors for high-quality
human synthesis, namely, dataset size, data distribution, and data alignment. 
% Furthermore, some advanced works~\cite{noguchi2022unsupervised,zhang2022avatargen,bergman2022gnarf,EVA3D} start to extending the generative capability of StyleGAN to 3D settings, in which neural rendering~\cite{mildenhall2022nerf,orel2022stylesdf} is introduced into the generator to synthesize 3D aware multi-view consistent human images. 
In this paper, we focus on the image-based VTON, which aims to generate realistic human image via fitting an in-shop garment image onto a reference person.

\commentH{Some of the mentioned works e.g. 3D neural rendering seems unrelated to our method. The relationship between our approach and these works is not covered as well.**Uncertain, to be confirmed.}

\begin{figure*}
  \centering
  \includegraphics[width=1.0\hsize]{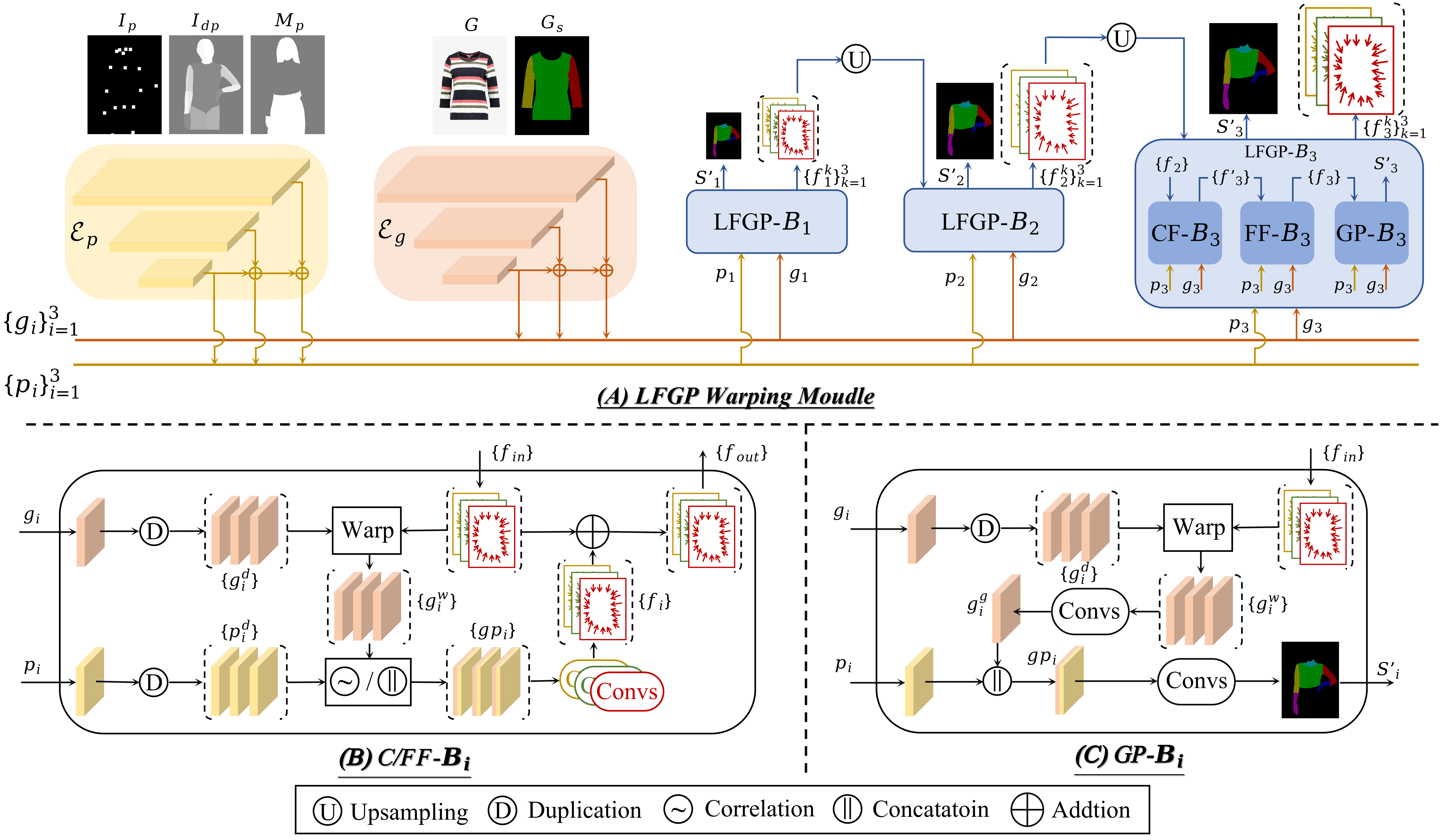}
  \vspace{-6mm}
  \caption{Overview of the LFGP warping module.\commentH{Should be $\{g_i\}^{N=3}_{i=1}$ here?}} 
  \vspace{-4mm}
  \label{fig:lfgp}
\end{figure*}

\textbf{Image-based Virtual Try-on.}
Most of the existing image-based VTON methods~\cite{bochao2018cpvton,xintong2019clothflow,ruiyun2019vtnfp,dong2019mgvton,han2020acgpn,thibaut2020wuvton,Ge2021PFAFN,he2022fs_vton,choi2021vtonhd,kedan2021ovnet,chopra2021zflow,lee2022hrviton,bai2022sdafn,morelli2022dresscode} follow a two-stage generation framework that separately deforms the in-shop garment to target shape and synthesizes the try-on result via combining the deformed garment and the reference person. Since the quality of the garment deformation directly determines the realism of the generated results, it is crucial to design a powerful deformation module in this generation framework. Some previous methods~\cite{bochao2018cpvton,ruiyun2019vtnfp,dong2019mgvton,han2020acgpn,thibaut2020wuvton,choi2021vtonhd} leverage neural network to regress sparse garment control points in target image, which are then used to fit a TPS transformation~\cite{bookstein1989tps} for garment deformation. 
% The TPS-based deformation is less expressive for the non-rigid transformation, due to the sparse control points. 
Other methods~\cite{xintong2019clothflow,Ge2021PFAFN,chopra2021zflow,he2022fs_vton,bai2022sdafn} instead estimate an appearance flow map~\cite{TinghuiZhou2016ViewSB} to model non-rigid deformation, where the flow map depicts the corresponding location in the source image for each pixel in the target image. 
Compared with TPS-based methods which fit a transformation function via the sparse correspondence between control points, the flow-based methods directly predict the dense correspondence for each pixel, thus is more expressive for complex deformation. However, both of the existing TPS- and flow-based methods directly learn a global deformation field for various garment parts, which is unable to model diverse local transformation for different garment parts. Therefore, they fail to obtain realistic deformed results when receiving the intricate human pose. In this paper, we innovatively learn diverse local deformation fields for different garment parts, thus is capable of handling challenging inputs. 
% It is worth noting that\commentH{It is worth noting that}, though SDAFN~\cite{bai2022sdafn} also learns multiple deformation filed for garment deformation, each deformation field is designed for the intact garment, thus it still fails to model part-aware deformation. 
Furthermore, exiting methods usually neglect the texture distortion around the protected region. To address this problem, we propose a dynamic gradient truncation strategy for network training. \commentH{Is it necessary to add that existing methods are not designed for general purpose?}

\section{Methodology}
Image-based virtual try-on algorithm aims to seamlessly transfer an in-shop garment $G$ onto a specific person $I$.\xd{emphasize which part is key factor to achieve general-purposed}
To achieve this, our GP-VTON proposes a Local-Flow Global-Parsing (LFGP) warping module (Sec.~\ref{subsection:lfgp}) to warp the garment to target shape, which first deforms local garment parts ${\{G^k\}}_{k=1}^3$ individually and then assembles different warped parts ${\{{G'}^k\}}_{k=1}^3$ together to obtain an intact warped garment $G'$. Besides, to address the texture distortion problem, GP-VTON introduces a Dynamic Gradient Truncation (DGT) training strategy (Sec.~\ref{subserction:dgt}) for the warping network. Finally, GP-VTON employs a try-on generator (Sec.~\ref{subsection:generator}) to synthesize the try-on result $I'$ according to $G'$ and other person-related inputs. Furthermore, GP-VTON can be easily extended for multi-category scenario and jointly trained by using data from various categories (Sec.~\ref{subsection:mcvton}). 
% The loss functions are provided in Sec.~\ref{subsection:loss}. 
An overview of GP-VTON is displayed in Fig.~\ref{fig:framework}.

\subsection{Local-Flow Global-Parsing Warping Module}\label{subsection:lfgp}
% As shown in Fig.~\ref{fig:lfgp}, 
% our LFGP warping module follows the flow estimation pipeline in~\cite{xintong2019clothflow,Ge2021PFAFN,he2022fs_vton,lee2022hrviton} and is composed of the pyramid feature extraction and the cascade flow estimation. In contrast to existing methods, we improve the flow estimation by using dedicatedly designed LFGP blocks to collaboratively estimate the diverse local flows and global garment parsing, which will be explained in details bellow.
As shown in Fig.~\ref{fig:lfgp}, 
our LFGP warping module follows the flow estimation pipeline in~\cite{xintong2019clothflow,Ge2021PFAFN,he2022fs_vton,lee2022hrviton} and is composed of pyramid feature extraction and cascade flow estimation. We will explain our dedicated improvement bellow.
% In contrast to existing methods, we improve the flow estimation by using dedicatedly designed LFGP blocks to collaboratively estimate the diverse local flows and global garment parsing, which will be explained in details bellow.

\textbf{Pyramid Feature Extraction.} 
Our LFGP warping module employs two Feature Pyramid Network (FPN)~\cite{Lin_2017_CVPR} (i.e., $\mathcal{E}_p$ and $\mathcal{E}_g$ in Fig.~\ref{fig:lfgp}) to separately extract the multi-scale person feature ${\{p_i\}}_{i=1}^N$ and  garment feature ${\{g_i\}}_{i=1}^N$. Specifically, $\mathcal{E}_p$ takes as inputs the human pose $I_p$, densepose pose $I_{dp}$, and the preserve region mask $M_p$, in which $I_p$ and $I_{dp}$ jointly provide the human pose information for flow estimation, and $M_p$ is essential for the generation of the preserved-region-aware parsing.  $\mathcal{E}_g$ takes as inputs the intact in-shop garment $G$ and its corresponding parsing map $G_s$, in which $G_s$ can explicitly provide the semantic information of different garment parts for parsing generation.\commentZ{Remember to explain how we obtain these conditions in the experiment section.} It is worth noting that, we extract five multi-scale features in our model (i.e., $N=5$) but set $N=3$ in Fig.~\ref{fig:lfgp} for brevity.

\textbf{Cascade Local-Flow Global-Parsing Estimation.} 
% Most of the existing methods~\cite{xintong2019clothflow,Ge2021PFAFN,he2022fs_vton,lee2022hrviton} directly leverage a global flow to warp the intact garment 
% % to target shape. 
% However, since the flow estimation network is trained in a weak supervised manner (i.e., without ground truth flow for supervision), it fails to model anisotropic deformation within a single flow map.\commentZ{Remember to modify the similar description in Intro. (i.e., trained in a weak supervised manner rather than unsupervised manner.)} 
% Therefore, when the warping module receives an intricate human pose and different garment parts require diverse deformation, 
% the global warping mechanism fails to estimate reasonable flow and tends to generate the unrealistic warped result.
Most of the existing methods~\cite{xintong2019clothflow,Ge2021PFAFN,he2022fs_vton,lee2022hrviton} directly leverage a global flow to warp the intact garment, which tends to generate unrealistic warped result 
% when the warping module receives an intricate human pose and different garment parts require diverse deformation.
when different garment parts require diverse deformation.
% to target shape. 
To solve this problem, our LFGP module explicitly divides the intact garment into three local parts, (i.e., left/right sleeve,  and torso region), and estimates three local flows to warp different parts individually.
% separately for different garment parts.
Since the deformation diversity within the same part is slight\commentH{little diversity, limited diversity confined diversity may be better?}, the local flow can handle warping precisely and produce semantic-correct warped result.
Furthermore, our LFGP estimates a global garment parsing to assemble local parts into an intact garment.

To be specific, as show in Fig.~\ref{fig:lfgp} (A), LFGP warping module exploits $N$ LFGP blocks to cascadingly estimate $N$ multi-scale local flows ${\{{\{f_i^k\}}_{k=1}^3\}}_{i=1}^N$ and global garment parsing ${\{S'_i\}}_{i=1}^N$.
% in which the local flows estimated in the current block will be refined in next block. 
Each LFGP block is composed of Coarse/Fine Flow Block (C/FF-B) and a Garment Parsing Block (GP-B), which estimate coarse/fine local flows and global garment parsing, respectively. As depicted in Fig.~\ref{fig:lfgp} (B), CF-B first duplicates the garment feature $g_i$ and employs the incoming local flows ${\{f_{in}\}}$, which comes from previous LFGP block, to warp the duplicated garment feature ${\{g_i^d\}}$ to three part-aware local warped features ${\{g_i^w\}}$. Then, the correlation operator from flownet2~\cite{IMKDB17} is employed to integrate the ${\{g_i^w\}}$ and duplicated person feature ${\{p_i^d\}}$ into three local fused features ${\{gp_i\}}$, which are separately sent into three convolution layers to estimate the corresponding local flows $\{f'\}$. At last, ${\{f'\}}$ is added to $\{f_{in}\}$ and produce the refined local flows $\{f_{out}\}$, which are the outputs of CF-B. FF-B has the same architecture as CF-F except 
it regards the output of CF-B as ${\{f_{in}\}}$ and directly concatenate ${\{g_i^w\}}$ and ${\{p_i^d\}}$ to obtain ${\{gp_i\}}$. 
For GP-B, as shown in Fig.~\ref{fig:lfgp} (C), it employs the refined local flows $\{f_{in}\}$ from FF-B to warp the duplicated feature ${\{g_i^d\}}$ to the part-aware local features ${\{g_i^w\}}$, which are then fused by convolution layers and becomes a global warped feature $g_i^g$. 
Finally, the concatenation of $g_i^g$ and the incoming $p_i$ is passed to convolution layers to estimate the global garment parsing $S'_i$,
whose labels consist of background, left/right sleeve, torso, left/right arms, and neck. % which is the output of GP-B. 
With the strong guidance of the warped feature $g_i^g$, GP-B is prone to generate garment parsing that the garment shape in different local region is consistent with its corresponding local warped part.

After finishing the last estimation in LFGP module, as displayed in Fig.~\ref{fig:framework} , GP-VTON deforms the local parts $\{G^k\}_{k=1}^3$ individually via their corresponding local flows ${\{f^k\}}_{k=1}^3$, and assembles local warped parts to an intact warped garment $G'$ by using the global garment parsing $S'$.

It is worth noting that, the global garment parsing is crucial for our local warping mechanism.
Since there will be overlap among different warped parts, directly assembling the warped parts together will leads to distinct artifact in the overlap region.
Instead, with the guidance of the global garment parsing, each pixel in the intact warped garment should be derived from a particular warped part, therefore the overlap artifact can be completely eliminated.

\begin{figure}
  \centering
  \includegraphics[width=1.0\hsize]{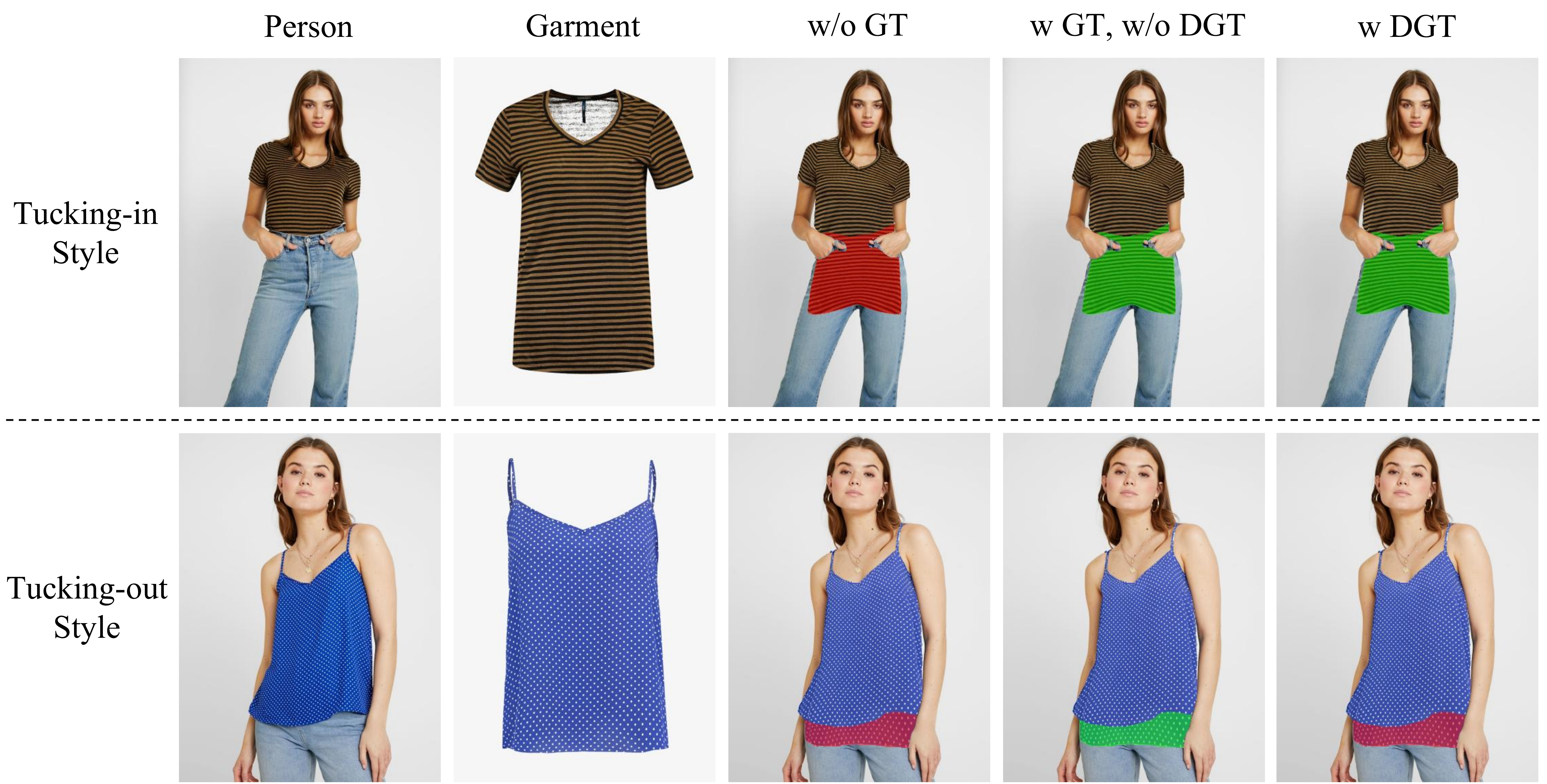}
  \vspace{-6mm}
  \caption{Comparison of different training strategies (i.e., gradient truncation (GT) and dynamic gradient truncation (DGT)) for LFGP. In the red region, the warped result will be penalized and the gradient will be backprograted. In the green region, the warped result will be neglected and the gradient will be truncated.} 
  % Please zoom in for more details.} 
  \vspace{-6mm}
  \label{fig:dgt}
\end{figure}

\subsection{Dynamic Gradient Truncation}\label{subserction:dgt}
Existing methods~\cite{bochao2018cpvton,han2020acgpn,Ge2021PFAFN,he2022fs_vton,lee2022hrviton} warp the in-shop garment according to the preserved region mask and force the warped garment to align with the boundary of the preserved region.
% (e.g., the junction of the upper and lower garment). 
However, directly warping the garment to meet the boundary constraint will lead to texture squeezing around the preserved region when the input person is in tucking-in style (as shown in the 1st case of Fig.~\ref{fig:dgt}.)

An intuitive solution for this issue is 
using the preserved mask to process the warped garment before calculating the training loss. In this way, the gradient in the preserve region will be truncated and the warped garment is no longer required to align with the boundary. 
However, when the training data is in tucking-out style (as shown in the 2nd case of Fig.~\ref{fig:dgt}.), gradient truncation is not appropriate since the inaccurate warped result in preserved region will not be penalized, leading to a stretched warped result.

To address above problems, our DGT training strategy dynamically conducts gradient truncation for different training samples according to their wearing style (i.e., tucking-in or tucking-out), in which the wearing style is determined by the disparity of the height-width ratio between the flat garment and the real warped garment (extracted from the person image). Fig.~\ref{fig:dgt} provides an intuitive comparison among different training strategies.
Specifically, we first extract the bounding boxes for the torso region of the flat garment and the warped garment by using their corresponding garment parsing. Then, we separately calculate the height-width ratio for each bounding box and use the ratio $R_{style}$ between the warped garment item $R^{warped}$ and flat garment item $R^{flat}$ to reflect the wearing style of current training sample, which can be formulated as:
\begin{equation}
R_{style} = R^{warped} / R^{flat},
\end{equation}
% \begin{equation}
% R^{*} = H^*/W^*,
% \end{equation}
where $R^{*} = H^*/W^*$, and $H^*$ and $W^*$ represent the height and width of the bounding box, respectively. 
We empirically find that if the person image is in tucking-in style, $R_{style}$ is usually less than 0.9, and if it is in tucking-out style, $R_{style}$ is usually more than 0.95. Therefore, we adopt gradient truncation when $R_{style}$ less than 0.9 and abolish it when $R_{style}$ more than 0.95. For training sample with $R_{style}$ between 0.9 and 0.95, we randomly adopt gradient truncation with a probability of 0.5.

\subsection{Try-on Generator}\label{subsection:generator}
After the flow-parsing estimation stage, GP-VTON employs a Res-UNet-based~\cite{HP_ronneberger2015u} generator $\mathcal{G}$ to synthesize the try-on result $I'$. As shown in Fig.~\ref{fig:framework}, 
% to generate $I'$,
$\mathcal{G}$ takes as inputs a skin color map $I_s$, a skin parsing map $S_s'$, a warped garment $G'$, and an image of  preserved region $I_p$, in which $I_s$ is a three-channel RGB image with the median value of the skin region (i.e., face, neck, arms.) while $S_s'$ is a one-channel label map that contains the skin region (i.e., neck, arms) in $S'$. 
% To obtain the final result $I'$, $\mathcal{G}$ first predicts a coarse try-on result $I'_c$ and an alpha mask $M_c$, then $M_c$ is used to fuse $I'_c$ and $G'$,  which can be formulated as:
% \begin{equation}
% I' = G' \odot M_c + I'_c \odot  (1-M_c).
% \end{equation}

\begin{figure}
  \centering
  \includegraphics[width=1.0\hsize]{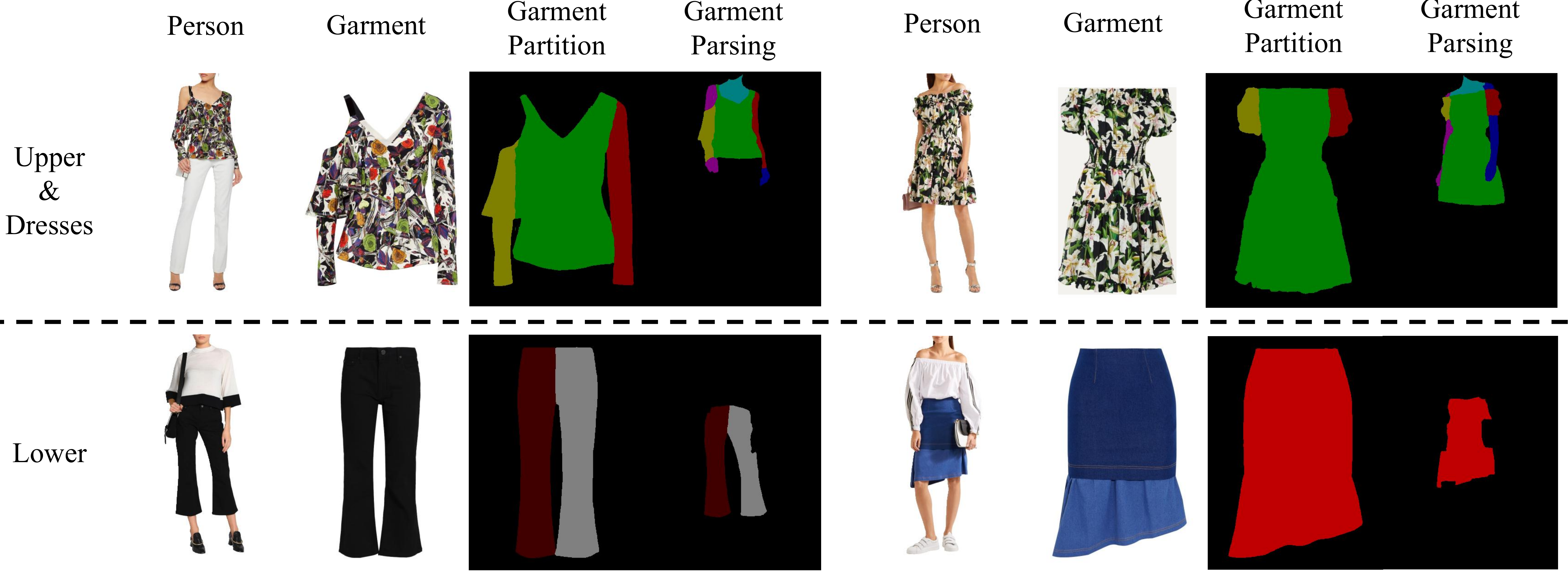}
  \vspace{-6mm}
  \caption{Illustration of the garment partition and the ground truth of garment parsing for different categories.} 
  \vspace{-6mm}
  \label{fig:garment_partition}
\end{figure}

\subsection{Multi-category Virtual Try-on}\label{subsection:mcvton}
Our GP-VTON can be easily extended to multi-category scenario through slight modifications. The core idea is using a unified partition mechanism for various garment categories. As mentioned in Dresscode~\cite{morelli2022dresscode}, common garments can be classified into three macro-categories (i.e., upper, lower, and dresses). Since upper garment and dresses have the similar topology, we can apply the same partition mechanism for upper garment and dresses, namely, dividing the garment into the left/right part (i.e., left/right sleeve) and the middle part (i.e., torso region). 
To make lower garment consistent with the other categories, we regard the pants and skirt as a single type and divide it into three parts, which is also composed of left/right-part (i.e., left/right pant leg), and middle-part (i.e., skirt).
By using this partition mechanism, garment from arbitrary categories can be divided into three local parts, which will be deformed individually and assembled to intact warped garment by our LFGP warping module.
% according to the estimated locals flow and global garment parsing, respectively. 
Besides, in the multi-category scenario, the estimated garment parsing is extended to include the labels of lower garment, which contain left/right pants, and the skirt. Fig.~\ref{fig:garment_partition} displays the garment partition and the ground truth garment parsing for different garment categories.

\subsection{Object Functions}\label{subsection:loss}
During training, we train LFGP warping module and the generator separately. For LFGP, 
we utilize $l_1$ loss $\mathcal{L}_1$ and perceptual loss~\cite{johnson2016perceptual} $\mathcal{L}_{per}$ for the warped results, and use $l_1$ loss $\mathcal{L}_{m}$ for the warped masks. We also use the pixel-wise cross-entropy loss $\mathcal{L}_{ce}$ and the adversarial loss $\mathcal{L}_{adv}$ for the estimated parsing.
% we calculate the $l_1$ loss  $\mathcal{L}_1$ and perceptual loss~\cite{johnson2016perceptual} $\mathcal{L}_{per}$ between local warped parts $\{G'^k\}_{k=1}^3$ and their corresponding ground truth $\{G_{gt}^k\}_{k=1}^3$. We also utilize the $l_1$ loss $\mathcal{L}_{m}$ for the local warped masks and pixel-wise cross-entropy loss $\mathcal{L}_{ce}$ for the global garment parsing. 
Besides, we follow PFAFN~\cite{Ge2021PFAFN} and employ the second-order smooth loss $\mathcal{L}_{sec}$ for the estimated flow. The total loss for LFGP module can be formulated as:
\begin{equation}
    \begin{split}
    \mathcal{L}^{w} = &\mathcal{L}_1^w + \lambda_{per}^w\mathcal{L}_{per}^w + \lambda_{m}^w\mathcal{L}_{m}^w \\
    &+\lambda_{ce}\mathcal{L}_{ce} + \lambda_{adv}^w\mathcal{L}_{adv}^w +
    \lambda_{sec}\mathcal{L}_{sec}.
    \end{split}
\end{equation}
% where $\lambda_{per}^w$, $\lambda_{m}^w$, $\lambda_{ce}$, and $\lambda_{sec}$ are the trade-off hyper-parameters. 
For the generator, we utilize $l_1$ loss $\mathcal{L}_1$, the perceptual loss~\cite{johnson2016perceptual} $\mathcal{L}_{per}$, and the adversarial loss for the try-on result $I'$, and also utilize the $l_1$ loss $\mathcal{L}_{m}$ for the alpha mask $M_c$. The total loss is defined as follows:
\begin{equation}
    \mathcal{L}^{g} = \mathcal{L}_1^g + \lambda_{per}^g\mathcal{L}_{per}^g + \lambda_{adv}\mathcal{L}_{adv} +
    \lambda_{m}^g\mathcal{L}_{m}^g.
\end{equation}
More details are provided in the supplementary materials. 
% where $\lambda_{per}^g$, $\lambda_{m}^g$, and $\lambda_{adv}$ are the hyper-parameters.

% \begin{equation}
% \mathcal{L}_{1} = \sum_{k=1}^{3}\|G'^k - G_{gt}^k \|_1,
% \end{equation}
% \begin{equation}
% \mathcal{L}_{perc} = \sum_{k=1}^{3}\sum_{j=1}^{5} \lambda_{j}\left\|\phi_{j}(G'^k)-\phi_{j}\left(G_{gt}^k \right)\right\|_{1},
% \end{equation}
% where $\phi_j(*)$ denotes the $j$-th feature map in a pre-trained VGG network~\cite{DBLP:journals/corr/SimonyanZ14a}.

% training details, first train without gradient truncation, then finetune with it

\begin{table}[t]

\def\arraystretch{1.2}
\small
\tabcolsep 2pt

\centering
\begin{tabular}{l c c c c c c c}
  \toprule
  \multicolumn{2}{c}{Method}                              
  & & SSIM $\uparrow$ & FID $\downarrow$ & LPIPS $\downarrow$ & mIoU $\uparrow$ & HE $\uparrow$ \\
  \cmidrule{1-2} \cmidrule{4-8} 
%   \multicolumn{2}{c}{CP-VTON~\cite{bochao2018cpvton}} 
%   & & 0.8615 & 15.54 & 0.1341 & ??? & ??? \\
%   \multicolumn{2}{c}{ACGPN~\cite{han2020acgpn}} 
%   & & ??? & ??? & ??? & ??? & ??? \\
  \multicolumn{2}{c}{PF-AFN~\cite{Ge2021PFAFN}} 
  & & 0.8858 & 9.475 & 0.0871 & 0.8412 & 14.9\% \\
  \multicolumn{2}{c}{FS-VTON~\cite{he2022fs_vton}} 
  & & 0.8829 & 9.552 & 0.0906 & 0.8357 & 8.80\% \\
  \multicolumn{2}{c}{HR-VITON~\cite{lee2022hrviton}} 
  & & 0.8623 & 16.21 & 0.1094 & 0.6949 & 9.10\% \\
  \multicolumn{2}{c}{SDAFN~\cite{bai2022sdafn}} 
  & & 0.8821 & 9.400 & 0.0922 & 0.5927 & 16.3\% \\
  \cmidrule{1-2} \cmidrule{4-8}
  \multicolumn{2}{c}{\textbf{GP-VTON (Ours)}} 
  & & \textbf{0.8939} & \textbf{9.197} & \textbf{0.0799} & \textbf{0.8764} & \textbf{50.9\%} \\  
  \bottomrule
\end{tabular}
\vspace{-2mm}
\caption{Quantitative comparisons on VITON-HD dataset~\cite{choi2021vtonhd}}
\vspace{-5mm}
\label{tab:vitonhd_results}
\end{table}

\begin{table*}[t]

\def\arraystretch{1.2}
\scriptsize
\tabcolsep 3pt
% \tiny
% \tabcolsep 2.5pt

\centering
\begin{tabular}{c c c c c c c c c c c c c c c c c c c c}
  \toprule
  \multicolumn{2}{c}{Dataset}& & \multicolumn{5}{c}{DressCode-Upper} & & \multicolumn{5}{c}{DressCode-Lower} & & \multicolumn{5}{c}{DressCode-Dresses} \\
  \cmidrule{1-2} \cmidrule{4-8} \cmidrule{10-14} \cmidrule{16-20}
  \multicolumn{2}{c}{Method}                              
  & & SSIM $\uparrow$ & FID $\downarrow$ & LPIPS $\downarrow$ & mIoU $\uparrow$ & HE $\uparrow$
  & & SSIM $\uparrow$ & FID $\downarrow$ & LPIPS $\downarrow$ & mIoU $\uparrow$ & HE $\uparrow$
  & & SSIM $\uparrow$ & FID $\downarrow$ & LPIPS $\downarrow$ & mIoU $\uparrow$ & HE $\uparrow$ \\
  \cmidrule{1-2} \cmidrule{4-8} \cmidrule{10-14} \cmidrule{16-20}

  \multicolumn{2}{c}{PF-AFN~\cite{Ge2021PFAFN}} 
  & & 0.9454 & 14.32 & 0.0380 & 0.8392 & 14.0\%
  & & 0.9378 & 18.32 & 0.0445 & 0.9463 & 12.3\%
  & & 0.8869 & 13.59 & 0.0758 & 0.8743 & 15.0\% \\
  \multicolumn{2}{c}{FS-VTON~\cite{he2022fs_vton}} 
  & & 0.9457 & 13.16 & 0.0376 & 0.8381 & 5.33\%
  & & 0.9381 & 17.99 & 0.0438 & 0.9478 & 14.7\%
  & & \textbf{0.8876} & 13.87 & 0.0745 & 0.8760 & 8.33\% \\
  \multicolumn{2}{c}{HR-VITON~\cite{lee2022hrviton}} 
  & & 0.9252 & 16.86 & 0.0635 & 0.6660 & 3.00\%
  & & 0.9119 & 22.81 & 0.0811 & 0.8670 & 2.67\%
  & & 0.8642 & 16.12 & 0.1132 & 0.7209 & 2.33\% \\
  \multicolumn{2}{c}{SDAFN~\cite{bai2022sdafn}} 
  & & 0.9379 & 12.61 & 0.0484 & 0.5046 & 11.3\%
  & & 0.9317 & \textbf{16.05} & 0.0549 & 0.4543 & 13.3\%
  & & 0.8776 & \textbf{11.80} & 0.0852 & 0.5945 & 19.3\% \\
  \cmidrule{1-2} \cmidrule{4-8} \cmidrule{10-14} \cmidrule{16-20}
  \multicolumn{2}{c}{\textbf{GP-VTON (Ours)}} 
  & & \textbf{0.9479} & \textbf{11.98} & \textbf{0.0359} & \textbf{0.8766} & \textbf{66.3\%}
  & & \textbf{0.9405} & 16.07 & \textbf{0.0420} & \textbf{0.9601} & \textbf{57.0\%}
  & & 0.8866 & 12.26 & \textbf{0.0729} & \textbf{0.8951} & \textbf{55.0\%} \\  
  \bottomrule
\end{tabular}
\vspace{-2mm}
\caption{Quantitative comparisons on DressCode dataset~\cite{morelli2022dresscode}}
\vspace{-2mm}
\label{tab:dresscode_results}
\end{table*}

\begin{table}[t]

\def\arraystretch{1.2}
\small
\tabcolsep 2pt

\centering
\begin{tabular}{l c c c c c c c c c c}
  \toprule
  \multicolumn{2}{c}{Method}                              
  & & LF  & GT & DGT  & & SSIM $\uparrow$ & LPIPS $\downarrow$ & mIoU $\uparrow$ & $R_{diff}$ $\downarrow$ \\
  \cmidrule{1-2} \cmidrule{4-6} \cmidrule{8-11} 

  \multicolumn{2}{c}{LFGP $\dagger$} 
  & & \xmark & \xmark & \xmark  & & 0.9016 & 0.0950 & 0.8412 & 0.3058 \\
  
%   \multicolumn{2}{c}{LFGP $\dagger$} 
%   & & \xmark & \cmark & \xmark & & 0.8927 & 0.1278 & 0.8148 & 0.1655 \\ 

%   \cmidrule{1-2} \cmidrule{4-6} \cmidrule{8-11} 

  \multicolumn{2}{c}{LFGP $\star$} 
  & & \cmark & \xmark & \xmark  & & 0.9039 & 0.0911 & \textbf{0.8784} & 0.3003 \\
  
  \multicolumn{2}{c}{LFGP $\ast$} 
  & & \cmark & \cmark & \xmark  & & \textbf{0.9053} & 0.0900 & 0.8774 & 0.2409 \\  
  
  \cmidrule{1-2} \cmidrule{4-6} \cmidrule{8-11} 
  \multicolumn{2}{c}{LFGP} 
  & & \cmark & \xmark & \cmark  & & 0.9050 & \textbf{0.0884} & 0.8764 & \textbf{0.1655} \\ 
  \bottomrule
\end{tabular}
\vspace{-2mm}
\caption{Ablation study of the Local FLow (LF), Gradient Truncation (GT), and Dynamic Gradient Truncation (DGT) on the VITON-HD dataset~\cite{choi2021vtonhd}.}
\vspace{-6mm}
\label{tab:ablation}
\end{table}

\begin{figure*}
  \centering
  \includegraphics[width=1.0\hsize]{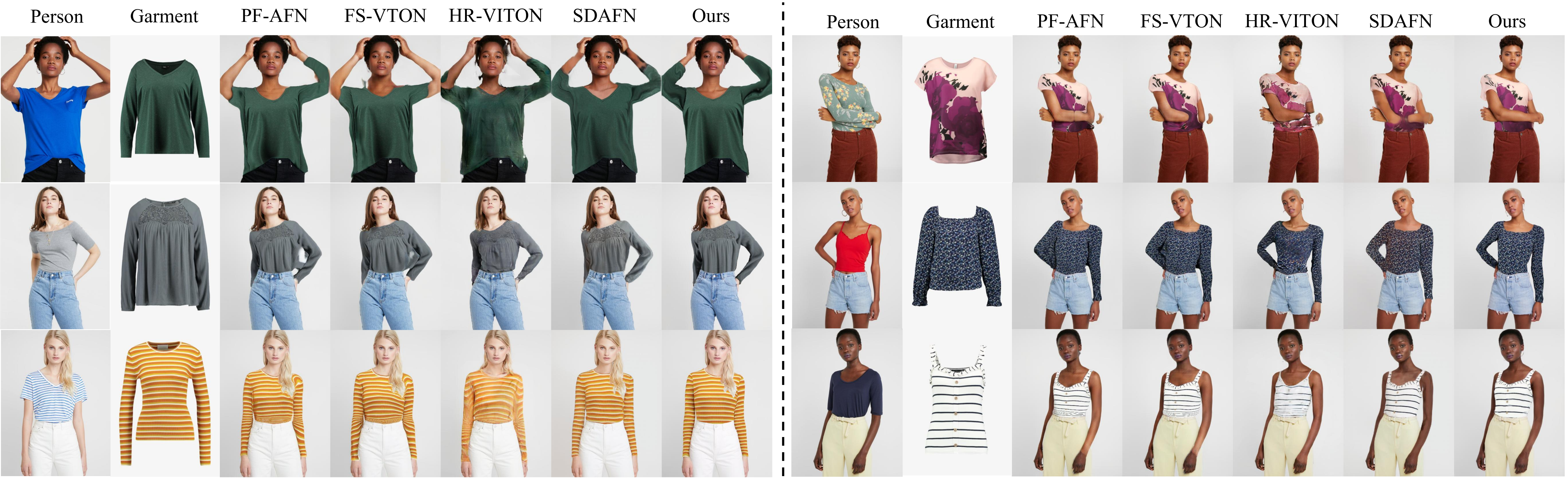}
  \vspace{-6mm}
  \caption{Qualitative comparison on VITON-HD dataset~\cite{choi2021vtonhd}.Please zoom in for more details.} 
  \vspace{-4mm}
  \label{fig:results_vitonhd}
\end{figure*}

\begin{figure*}
  \centering
  \includegraphics[width=1.0\hsize]{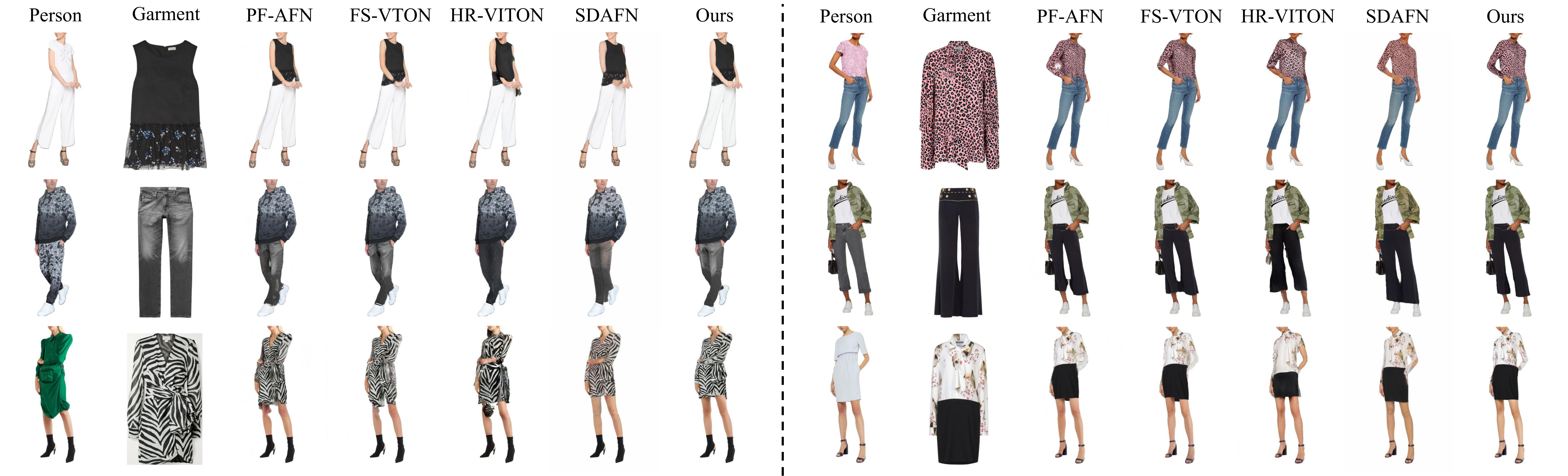}
  \vspace{-6mm}
  \caption{Qualitative comparison on Dresscode dataset~\cite{morelli2022dresscode}.Please zoom in for more details.} 
  \vspace{-5mm}
  \label{fig:results_dresscode}
\end{figure*}

\begin{figure}
  \centering
  \includegraphics[width=1.0\hsize]{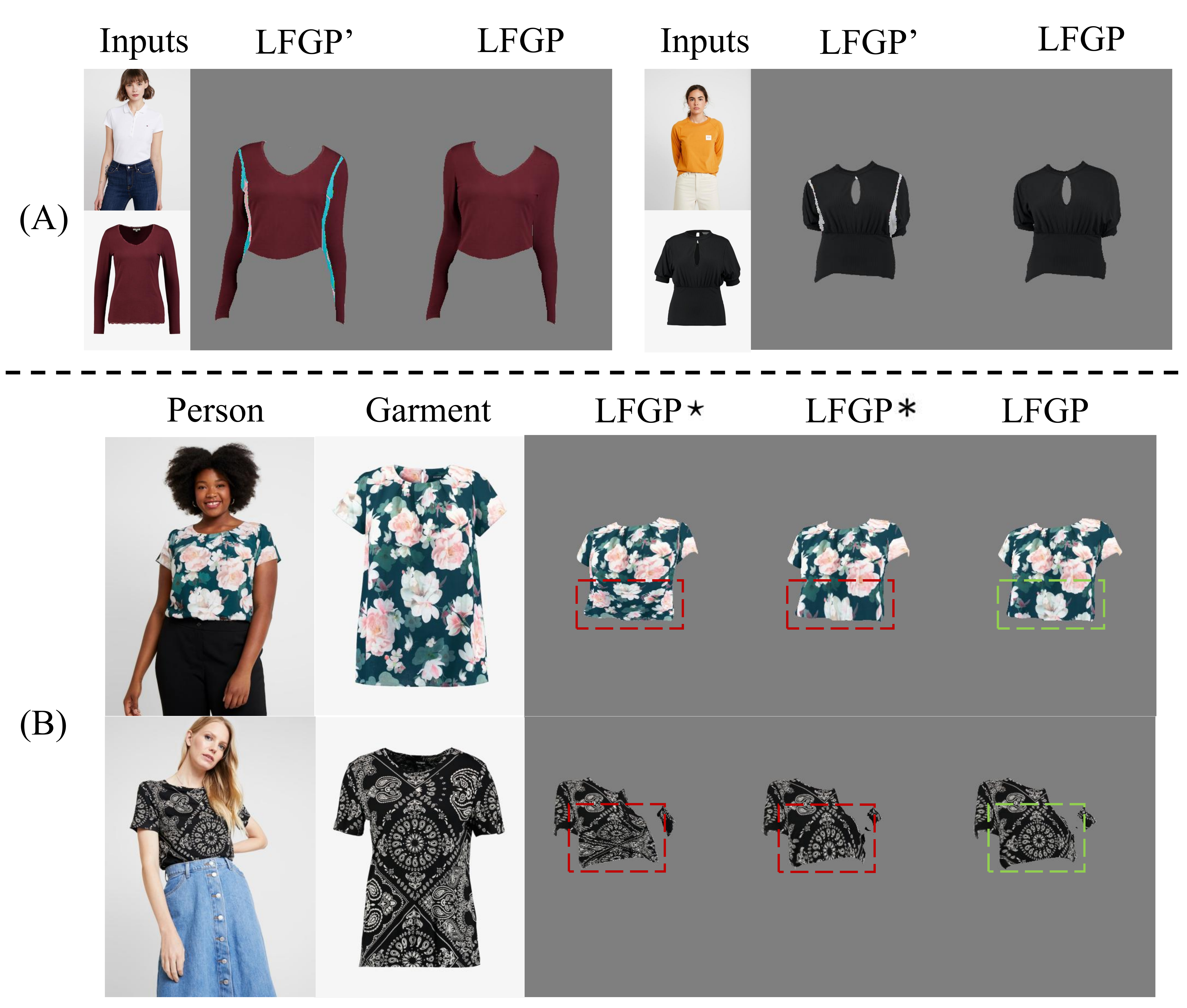}
  \vspace{-6mm}
  \caption{Ablation studies on the effectiveness of (A) the global parsing during the parts assembling process and (B) the dynamic gradient truncation training strategy.} 
  % Please zoom in for more details.} 
  \vspace{-5mm}
  \label{fig:ablation_study}
\end{figure}

\section{Experiments}

\noindent\textbf{Datasets.}
Our experiments are conducted under the resolution of $512 \times 384$ by using two existing high-resolution virtual try-on benchmarks VITON-HD~\cite{choi2021vtonhd} and DressCode~\cite{morelli2022dresscode}. VITON-HD contains 13,679 image pairs of front-view upper-body woman and upper garment, which are further split into 11,647/2,032 training/testing pairs. 
DressCode is composed of 48,392/5,400 training/testing pairs of front-view full-body person and garment from different categories (i.e., upper, lower, dresses). 
For each dataset, we employ~\cite{openpose} and~\cite{PE_facebook2018densepose} to extract the 2D pose and densepose, respectively. Besides, we apply a unified parsing estimator to predict the human/garment parsing for person/garment image, in which the estimator is based on ~\cite{cheng2021mask2former} and trained by using 80k manual annotated fashion images.
% We will release the estimated results for both dataset for the follow-up virtual try-on studies.
 
% \noindent\textbf{Implementation Details.}
% The training process for both dataset are the same, which include a two-stage training procedure and are trained on 8 Tesla V100 GPUs. During training LFGP warping module, the batch size is set to 2 for each GPU and the model is trained for 120 epochs with learning rate 5e-5, in which the DGT strategy is only employed for the last 50 epochs. During training the generator, the batch size is set to 16 for each GPU and the model is trained for 200 epochs with learning rate 5e-4. 
% More details about the network architecture, training objects, and hyperparameter setting are provided in the supplementary materials.

\noindent\textbf{Baselines and Evaluation Metrics.}
We compare GP-VTON with several stage-of-the-art methods, including PF-AFN~\cite{Ge2021PFAFN}, FS-VTON~\cite{he2022fs_vton}, HR-VITON~\cite{lee2022hrviton}, and SDAFN~\cite{bai2022sdafn}, which are trained from scratch on VITON-HD~\cite{choi2021vtonhd} and DressCode~\cite{morelli2022dresscode} through using the official codes provided by the authors.

We employ three widely used metrics (i.e., Structural SIMilarity index (SSIM)~\cite{Wang2004SSIM}, Perceptual distance (LPIPS)~\cite{zhang2018perceptual}, and Fr$\mathbf{\acute{e}}$chet Inception Distance (FID)~\cite{parmar2021cleanfid}) to evaluate the similarity between synthesized and real images, in which SSIM and LPIPS are used for paired setting and FID are used for unpaired setting. We also utilize the mean Intersection over Union (mIoU) between the warped garment parsing and its corresponding ground truth (extracted from the human parsing) to evaluate the semantic-correctness of the warping module in different methods. Furthermore, we conduct Human Evaluation (HE) to evaluate different methods according to their synthesis quality.\footnote{More details about architecture details, implementation details, and user study setting are provided in the supplementary materials.}

\subsection{Qualitative Results}
Fig.~\ref{fig:results_vitonhd} and Fig.~\ref{fig:results_dresscode} display the qualitative comparison of GP-VTON with the state-of-the-art baselines on VITON-HD dataset~\cite{choi2021vtonhd} and DressCode dataset~\cite{morelli2022dresscode}, respectively. 
% Compared with the baselines, GP-VTON is capable of generating more realistic try-on results. 
Both figures demonstrate the superiority of GP-VTON over the baselines.
First of all,  the baselines fail to generate semantic-correct try-on results when encountering intricate poses,  resulting in the damaged sleeves and arms (e.g., 1st row in Fig.~\ref{fig:results_vitonhd}), the blended pant legs and the indistinguishable sleeve (e.g., 1st case of 2nd row and 1st case of 3rd row in Fig.~\ref{fig:results_dresscode}). 
Second, when receiving a complex garment (i.e., without obevious interval between adjacent parts), the baselines are prone to generate adhesive artifact (e.g., 2nd row in Fig.~\ref{fig:results_vitonhd} and 2nd case of 2nd row in Fig.~\ref{fig:results_dresscode}).
Third, existing methods~\cite{Ge2021PFAFN,he2022fs_vton,lee2022hrviton} 
tend to generate distorted texture around the preserved region (e.g., the third row in Fig.~\ref{fig:results_vitonhd}).
In comparison, GP-VTON first employs local flows to warp different garment parts individually, leading to the precise local warped parts, and then uses the global garment parsing to assemble local parts into an semantic-correct warped garment. Therefore,  GP-VTON is more robust to intricate pose or complex input garment.
Besides, by using the dynamic gradient truncation training strategy, GP-VTON can avoid generating distorted texture around preserved region. 

\subsection{Quantitative Results}
As reported in Tab.~\ref{tab:vitonhd_results}, our GP-VTON consistently surpasses the baselines on all metrics for the VITON-HD dataset~\cite{choi2021vtonhd}, demonstrating that GP-VTON can obtain more precise warped garments and generate try-on results with better visual quality. Particularly, on the mIoU metric, GP-VTON outperforms other methods by a large margin, which further illustrates that our LFGP warping module is capable of obtaining the semantic-correct warped results.
Tab.~\ref{tab:dresscode_results} shows the quantitative comparisons of GP-VTON with other methods on the DressCode dataset~\cite{morelli2022dresscode}. As shown in the table, for DressCode-Upper, GP-VTON achieves the finest score on all metrics. For DressCode-Lower and DressCode-Dresses, GP-VTON outperforms other methods on most of the metrics and obtains comparable low FID score with SDAFN~\cite{bai2022sdafn}. This is mainly because the human pose in the DressCode-Lower and DressCode-Dresses are generally simple, which do not require complex warping during try-on process, thus the advanced SDAFN~\cite{bai2022sdafn} can also obtain compelling FID score. However, the superiority of GP-VTON on the mIoU and HE metrics can still indicate its warped results are more semantically correct and its synthesized results are more photo-realistic.

\subsection{Ablation Study}
To validate the effectiveness of LFGP warping module and DGT training strategy, we design three variants of our proposed method and evaluate the performance of different variants according to their metric scores for the warped results. Besides, we define another metric $R_{diff}$ to measure the difference of the height-width ratio between the warped garment and the in-shop garment, where the lower value indicates the better preservation of the original height-width ratio, thus implying the better warping.

We regard PF-AFN~\cite{Ge2021PFAFN} as our first variant (denoted as  LFGP$\dagger$), since it utilizes a global flow for warping and is trained without Gradient Truncation. We further implement the other two variants (i.e., LFGP$\star$ and LFGP$\ast$) by training LFGP module without Gradient Truncation and without Dynamic Gradient Truncation, respectively.

\noindent\textbf{LFGP Module.} As reported in Tab.~\ref{tab:ablation}, compared with LFGP$\dagger$, other methods with local flows gain increase on SSIM and LPIPS, and achieve obvious improvement on the mIoU metric, demonstrating that our local flows warping mechanism can obtain more realistic and semantic-correct warped results. Besides, we further conduct another experiment on the full LFGP model (denoted as LFGP') to demonstrate the  effectiveness of the global parsing. As shown in Fig.~\ref{fig:ablation_study} (A), by using the global parsing to assemble different warped parts, the overlap artifact between different parts can be completely eliminated.

\noindent\textbf{DGT Training Strategy.}
As reported in Tab.~\ref{tab:ablation}, compared with LFGP$\star$, LFGP$\ast$ with the normal GT strategy obtains lower $R_{diff}$ score while the full LFGP model with DGT strategy achieves the lowest $R_{diff}$ score.
Fig.~\ref{fig:ablation_study} (B) further provide the visual comparisons among different methods, in which LFGP$\star$ tends to squeeze the texture while LFGP$\ast$ tends to stretch the texture. In contrast, the full LFGP module can preserve the texture details well.
Both of the quantitative and qualitative comparisons validate that training with DGT facilitates the warping model to preserve the original height-width ratio of the garment, thus avoid texture squeezing or stretching.

\section{Conclusion}
In this work, we propose GP-VTON towards the general-purpose virtual try-on, which is capable of generating semantic-correct and photo-realistic try-on results even in the challenging self-occlusion scenarios and can be easily extended to multi-category scenarios. Specifically, to make garment warping robust to intricate inputs, GP-VTON introduces the Local-Flow Global-Parsing (LFGP) warping module to warp local parts individually and assembles local warped parts via the estimated global garment parsing. Besides, to alleviate the texture distortion problem in existing methods, GP-VTON employs a Dynamic Gradient Truncation (DGT) training strategy for the warping network. Experiments on two high resolution virtual try-on benchmark illustrate GP-VTON's superiority over existing methods.
The limitation and social impact of our GP-VTON will be discussed in the supplementary materials.

\section{Acknowledgement}
This work was supported in part by National Key R$\&$D Program of China under Grant No.2020AAA0109700, National Natural Science Foundation of China (NSFC) under Grant No.61976233, Guangdong Outstanding Youth Fund (Grant No.2021B1515020061), Shenzhen Fundamental Research Program (Project No.RCYX20200714114642083, No.JCYJ20190807154211365), the Fundamental Research Funds for the Central Universities, Sun Yat-sen University under Grant No.22lgqb38.

\clearpage

%%%%%%%%% REFERENCES
{\small
\bibliographystyle{ieee_fullname}
\bibliography{main}
}

\end{document}